\documentclass{article}

\usepackage{arxiv}

\usepackage[utf8]{inputenc} % allow utf-8 input
\usepackage[T1]{fontenc}    % use 8-bit T1 fonts
\usepackage{hyperref}       % hyperlinks
\usepackage{url}            % simple URL typesetting
\usepackage{amsmath}
\usepackage{booktabs}       % professional-quality tables
\usepackage{amsfonts}       % blackboard math symbols
\usepackage{nicefrac}       % compact symbols for 1/2, etc.
\usepackage{microtype}      % microtypography
\usepackage{lipsum}		% Can be removed after putting your text content
\usepackage{graphicx}
\graphicspath{ {./images/} }

\title{A Compressive Sensing Video dataset using Pixel-wise coded exposure}

\author{
  Sathyaprakash Narayanan \\
  Department of Electronics and Systems Engineering \\
  Indian Institue of Science \\
  Bangalore, India 560012 \\
  \texttt{sathyaprakas@iisc.ac.in} \\
   \And
 Yeshwanth Ravi Theja Bethi \\
Department of Electronics and Systems Engineering \\
  Indian Institue of Science \\
  Bangalore, India 560012 \\
  \texttt{yeshwanthb@iisc.ac.in} \\
  \AND
  Chetan Singh Thakur \\
  Assistant Professor \\
  Department of Electronics and Systems Engineering \\
  Indian Institue of Science \\
  Bangalore, India 560012 \\
  \texttt{csthakur@iisc.ac.in}
}

\begin{document}
\maketitle

\begin{abstract}
Manifold amount of video data gets generated every minute as we read this document, ranging from surveillance to broadcasting purposes. There are two roadblocks that restrain us from using this data as such, first being the storage which restricts us from only storing the information based on the hardware constraints. Secondly, the computation required to process this data is highly expensive which makes it infeasible to work on them. Compressive sensing(CS)\cite{baraniuk2017compressive} is a signal process technique\cite{erlich2009dna}, through optimization, the sparsity of a signal can be exploited to recover it from far fewer samples than required by the Shannon-Nyquist sampling theorem. There are two conditions under which recovery is possible. The first one is sparsity which requires the signal to be sparse in some domain. The second one is incoherence which is applied through the isometric property which is sufficient for sparse signals\cite{donoho2006most}\cite{davenport2013fundamentals}. To sustain these characteristics, preserving all attributes in the uncompressed domain would help any kind of in this field. However, existing dataset fallback in terms of continuous tracking of all the object present in the scene, very few video datasets have comprehensive continuous tracking of objects. To address these problems collectively, in this work we propose a new comprehensive video dataset, where the data is compressed using pixel-wise coded exposure \cite{hitomi2011video} that resolves various other impediments.
\end{abstract}

% keywords can be removed
\keywords{Video dataset \and Compressive Sensing \and Pixel wise coded exposure \and Object Recognition \and Tracking \and Reconstruction \and Localization }

\section{Introduction}
\begin{figure}[!h]
    \centering
    \includegraphics[width=\textwidth]{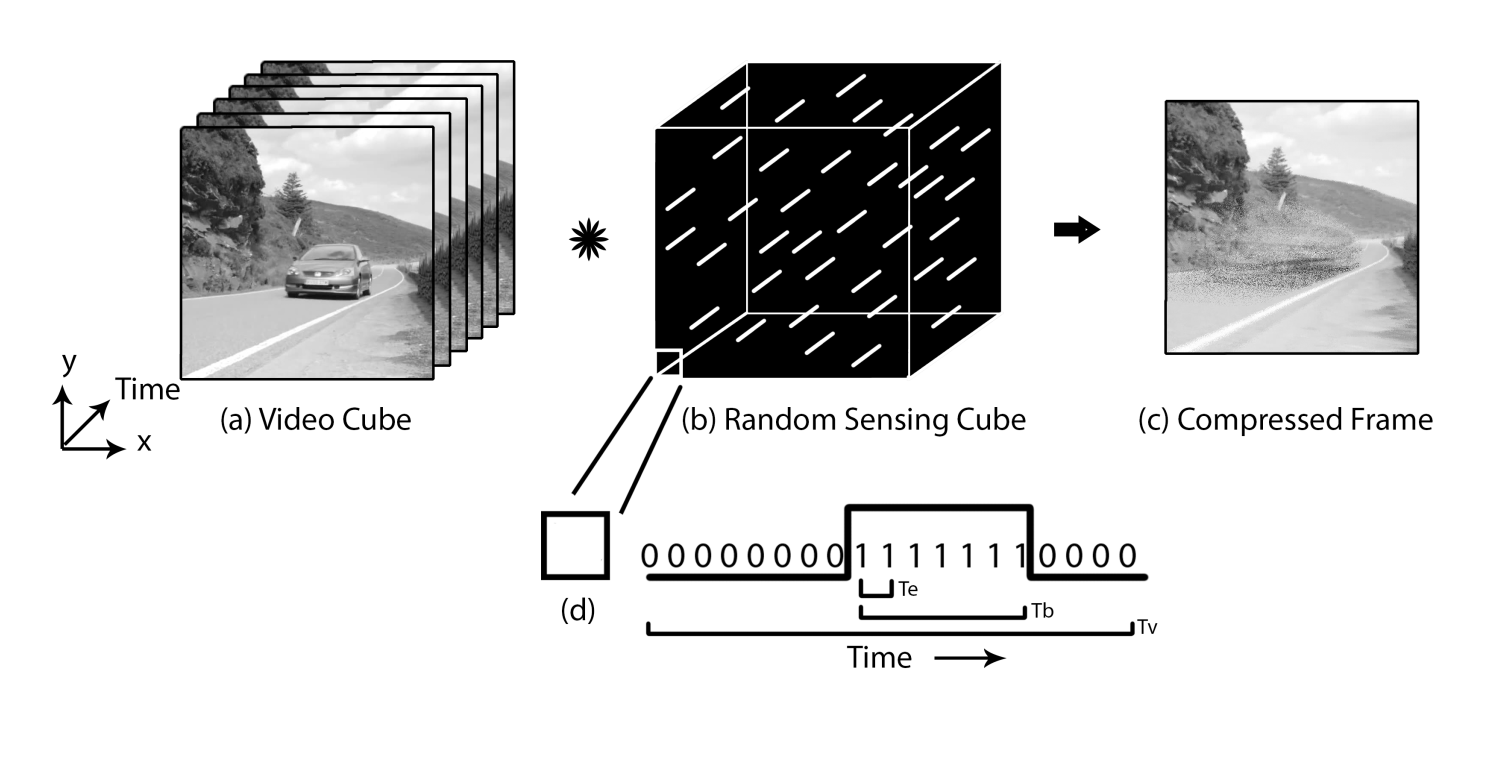}
\caption{Compression using pixel wise coded exposure}
  \label{fig:fig1}
\end{figure}

Surveillance videos are generated ubiquitously and having humans to monitor them constantly is quite a challenge. In most cases, a combination of an anomaly detection algorithm, object detection algorithm, etc.. are used to detect any aberration in these footage’s. The problem being that, each frame has to be processed and it requires an immense amount of memory and resources to store and process the footage’s. Also, the privacy has always been a constant fear, that hinders the usage of these raw footage's. For our rescue, we have the compressive sensing algorithm by which the amount of data to be processed is reduced by \textit{K (compression rate)} times, which not only reduces the space to store the data but in turn helps to expedite the process. A compressed frame Fig.\ref{fig:fig1}  is obtained by compressing \textit {K} number of frames along the temporal window by which it saves the memory for storing \textit {K-1} frames. Hence the computation is required only for \[
(N/K) data points
\begin{cases}   
\text{where,  N: Total data points} \\
\text{\hspace{10.5mm}K: Compression rate }                                                                                      \end{cases}
\] 
  
\begin{figure}[!h]
\centering
\includegraphics[width=\textwidth]{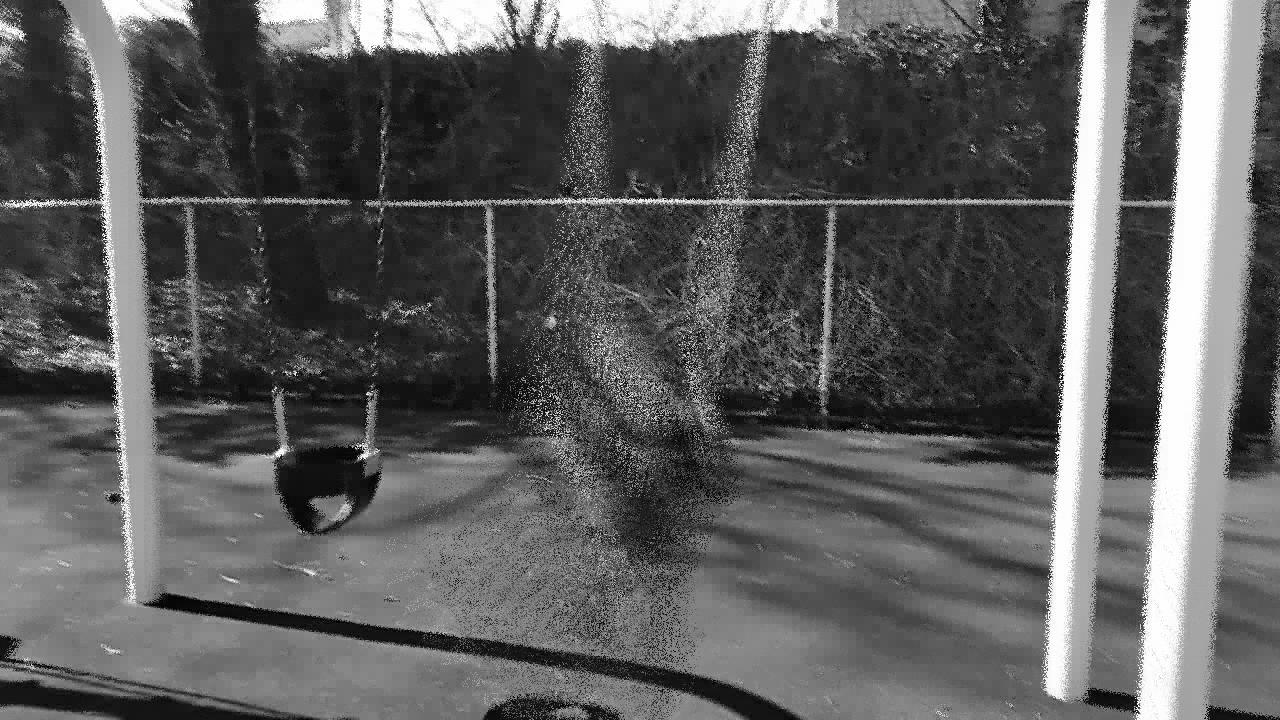}
\caption{Compressed frame}
\label{fig:fig2}
\end{figure}

In a compressed frame \ref{fig:fig2} the spatial information present is very hard to recondite even by a human or any conventional algorithm. This acts as an additional layer of encryption by encapsulating the moving information present in the data and allows only the owner to reconstruct back the whole scene.  We leverage on this this fact to convert the videos into the sparse domain and reduce the memory required for storage and the overall processing power for any any kind of signal processing directly on the compressed frames at the same moment converse the private information preset. We hope to embark a new stream of research on this sector.

\section{Challenges for the dataset}

\subsection{Video}
\begin{figure}[!h]
    \centering
    \includegraphics[width=\textwidth]{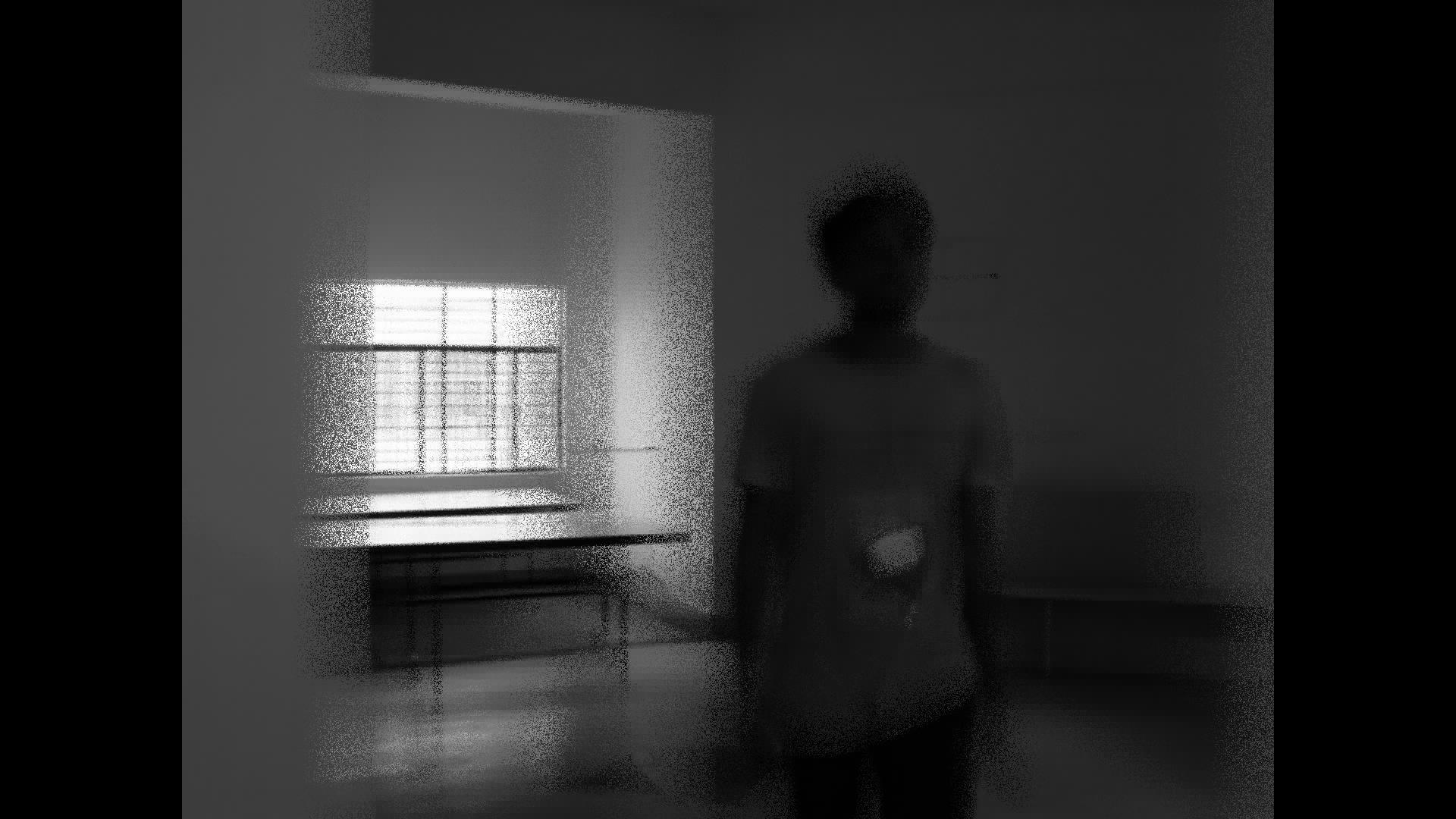}
\caption{ Motion blur as both the object and the camera is in motion}
\label{fig:fig3}
\end{figure}

\begin{enumerate}
 \item \textbf{Video Processing} \\
Since during the compression all the frames are to be in natural order without any kind of artificial transitions or angle cuts, since the compression would also consider the transition as a motion and would incorporate it for the compression. Hence, the dataset should also adhere to the fact that the video should not contain any artificial processing like angle cuts or transitions. \cite{baraniuk2017compressive}.
 \item \textbf{Motion and Environment} \\
 Since being the first dataset of its kind, we tend to capture all sorts of motions. That would fill the spectrum of motions ranging from moving object and stationary background, moving an object and changing background, stationary object and changing background, stationary object and stationary background Fig.\ref{fig:fig3}.
 \end{enumerate}
\subsection{Labelling} 
\begin{enumerate}
 \item Objects and Tracking  \\
 Since in compressive sensing all the objects in the set of \textit{K} frames are being compressed onto a single frame, the resultant tend to contain abstruse information in the compressed frame. The videos in the dataset should have the labels for each frame tracked along time (tracking) \cite{Wu_2013_CVPR}. There exist few related works in the same lines that contains the tracking information of certain objects. But it lacks the fact that all the objects present in the video are labeled as shown in Fig \ref{fig:fig7} and tracked along with the frames. As this is an essential feature for any kind of operation that is to be performed directly on the compressed frame, we have to store all the information of each frame of the video. 

\end{enumerate}

 Datasets on a collective note would be used in different countries for different intentions. Hence, the generalization of scenes from all parts of the world that could resemble a generic form should be considered. So as to incorporate some variance into the dataset.\\

 Thus to solve all the above challenges collectively and to savor the embodiment's available in this domain, we present out work that comprehensively pertains almost every aspect that is required for any kind of advancement in this field.
 
 \section{Dataset}
 
We searched the Youtube 8M\cite{abu2016youtube} dataset for videos which had only a single person or a single car. As all the variations were not available, we collected more videos of our own within our university campus for more training data. Since these are the preliminary stages of research we 

\begin{figure}[!h]
\centering
\includegraphics{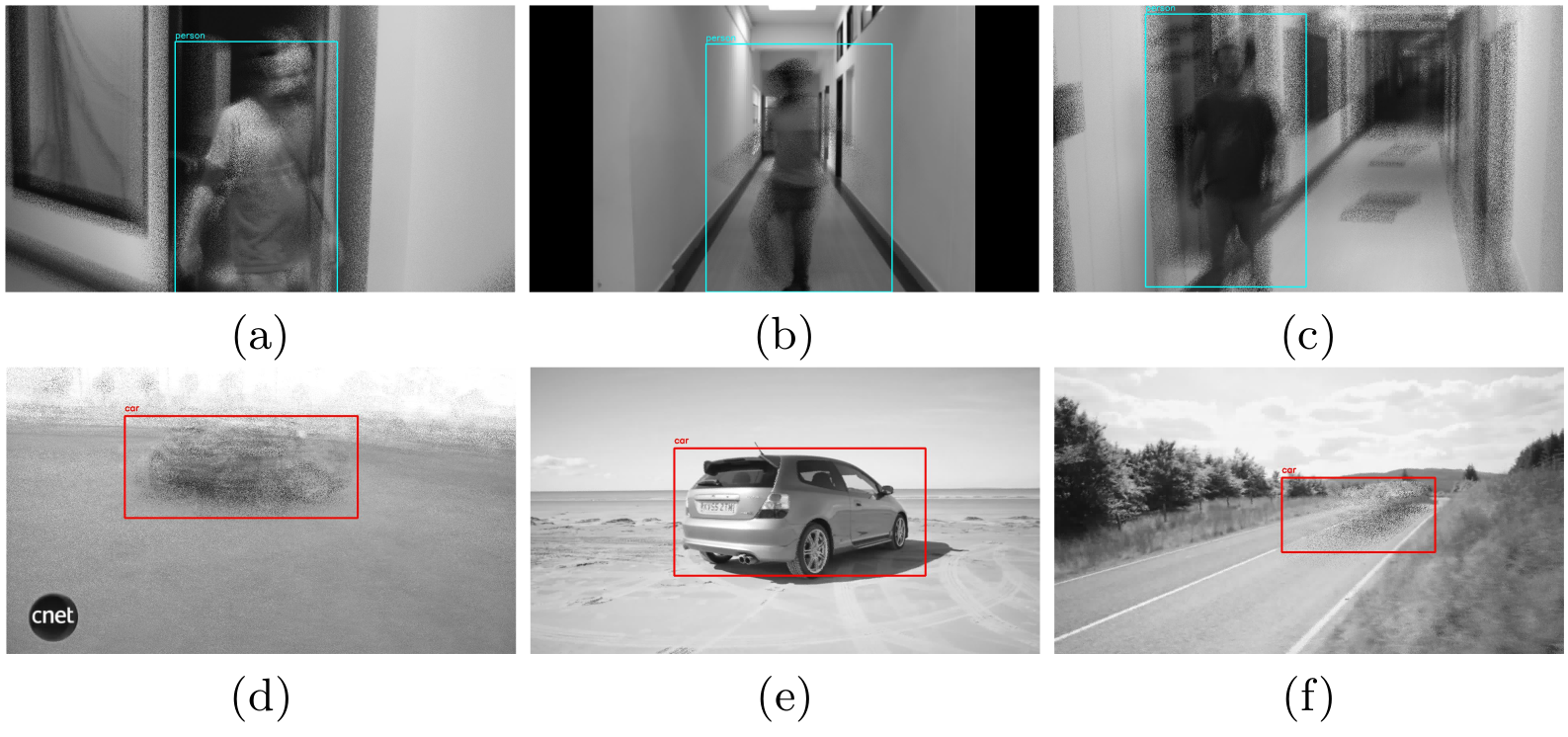} \\
\caption{(a)Both person and the background in motion (b)Static background and moving person (c)Stationary persona and moving background (d)Both car and the camera is motion (e)Both background and object stationary (f)Static background and moving car}
\label{fig:fig4}
\end{figure}

We made sure that only one object of interest was present in all of the videos captured. We captured videos on multiple days and during different parts of the day and tried to maintain as much variance as possible. The videos include objects moving with a stationary background, stationary objects with moving background and both the object and background moving \ref{fig:fig4}. The amount of movement in the frames were also varied. 

We chose Persons and Car as the only(currently) classes because it was difficult to collect data of other classes and also these two were the two most important object classes for surveillance tasks.

Hitomi et.al.,\cite{hitomi2011video} showed that compression rate $K$ between 9X-18X and bump time(Tb) 3-4 produces the best PSNR in the reconstruction of compressive sensed frames. We set the bump duration(Tb) to 3 frames and the compression to 13X for all the CS frames in the dataset as these would be more likely used in real-world cases. We also changed the sensing matrix while creating each CS frame, on speculation for research in this domain.

Being the first of its kind of dataset, we tried to provide the preliminary label attributes for the dataset. 
\begin{itemize}
\item Object class
\item Boundary box information \textit (tracked along the temporal window of $K$ frames)
\end{itemize}

\begin{enumerate}
 \item \textbf{Methodology} \\
 \begin{enumerate}
  \item \textbf{Compression Technique using PCE}   \\
We took videos with 30 FPS  and compressed them with 13X compression rate. The bump time Te is set to 3. For every 13 frames in the original video, a single compressed frame is generated using the above equation. The random sensing matrix has been changed for each set of 13 frames, so as to generalize compressed sensed frames and not to fit a particular sensing matrix. We normalize the pixel values after adding them through 13 frames to constrain the pixel value within 255 Fig.\ref{fig:fig3}. We only worked with monochrome videos because conventionally only monochromatic CMOS sensors \cite{xiong2017live} are available.

\begin{figure}[!h]
\centering
\includegraphics[scale=0.5]{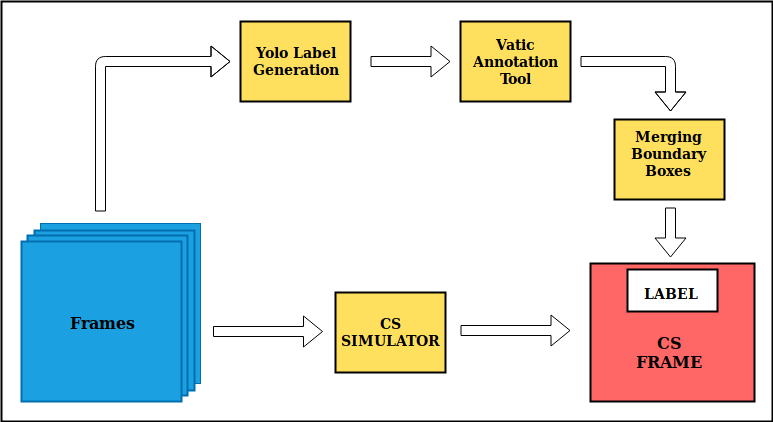}
\caption{Pipeline for compression}
\label{fig:fig5}
\end{figure}

\item \textbf{Psuedo labels for ground truth} \\
Since this was the first attempt to do object detection in CS frames, there was no dataset available publicly for this purpose, we had to collect our own training data. Manual labeling of the CS frames is a difficult job to do as the edges of moving objects often are indeterminable even to human eye. To avoid this problem, we chose pseudo-labeling of the data. As most of the existing architectures like YOLO v3\cite{redmon2018yolov3} detect persons and cars classes with really good accuracy, we used trained YOLO v3 to pseudo label the CS frames Fig.\ref{fig:fig7}. 
\begin{figure}[!h]
\centering
\includegraphics[width=\textwidth]{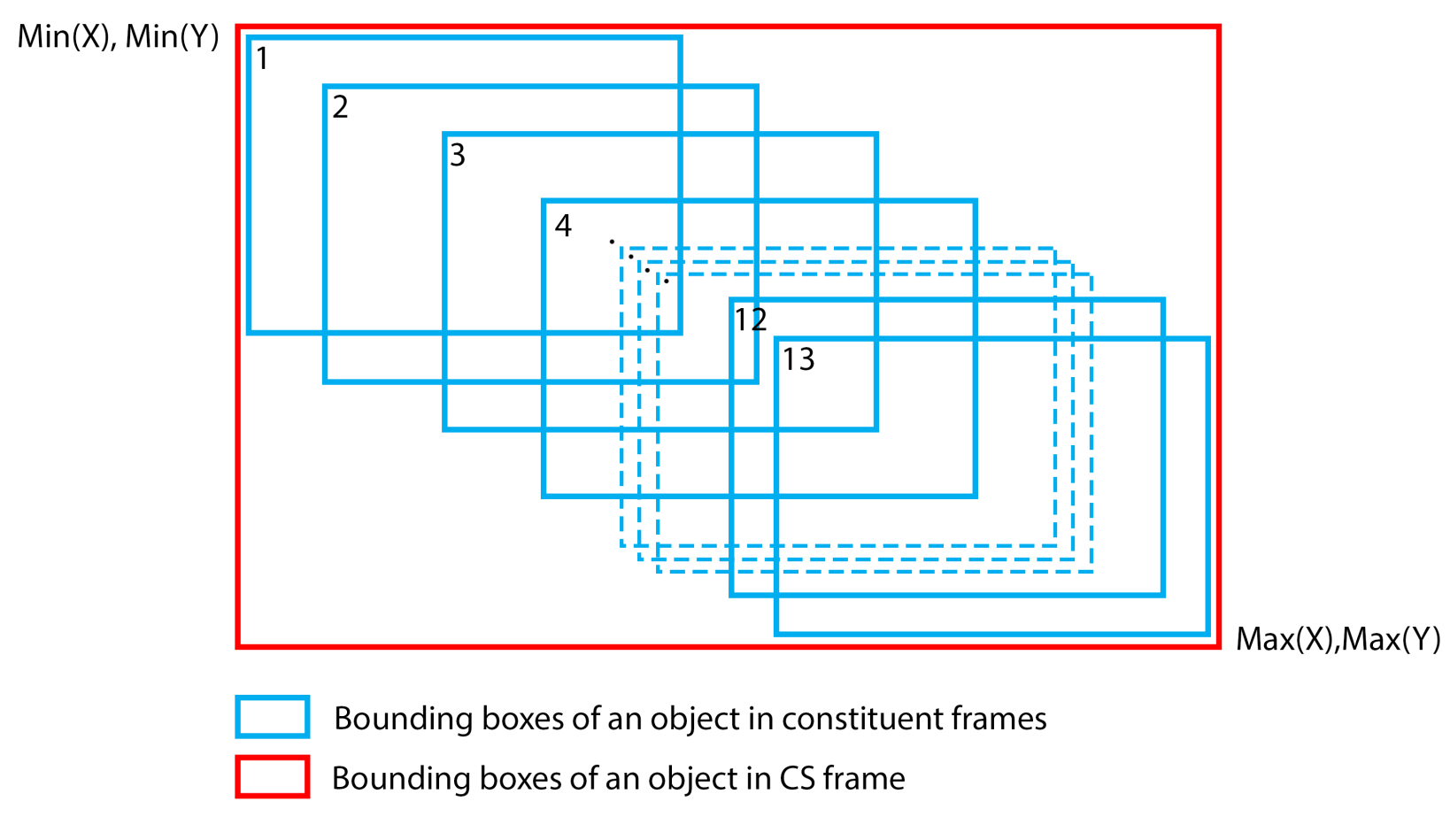}
\caption{Pseudo labelling}
\label{fig:fig6}
\end{figure}

We also had to come up with a way to draw bounding boxes in the CS domain. So, we chose to draw a bigger bounding box in the CS frame which enclosed all the individual bounding boxes of the constituent frames. Fig.\ref{fig:fig6} shows the method to merge the bounding boxes of each object across the constituent frames. We take the bounding box coordinates of an object in each frame, find the minimum and maximum of its X and Y coordinates and use them as the coordinates for the bigger bounding box in the CS frame. Tracking a single object's detection in consecutive frames which contain multiple objects of the same class is difficult because YOLO doesn't order them in any particular fashion. Because of this difficulty to attach each bounding box and track it along the frames to a specific object, for training dataset we collected videos which have only one object of interest i.e., frames having only one and the same car or person in each frame \ref{fig:fig7}.

\begin{figure}[!h]
\centering
\includegraphics[width=\textwidth]{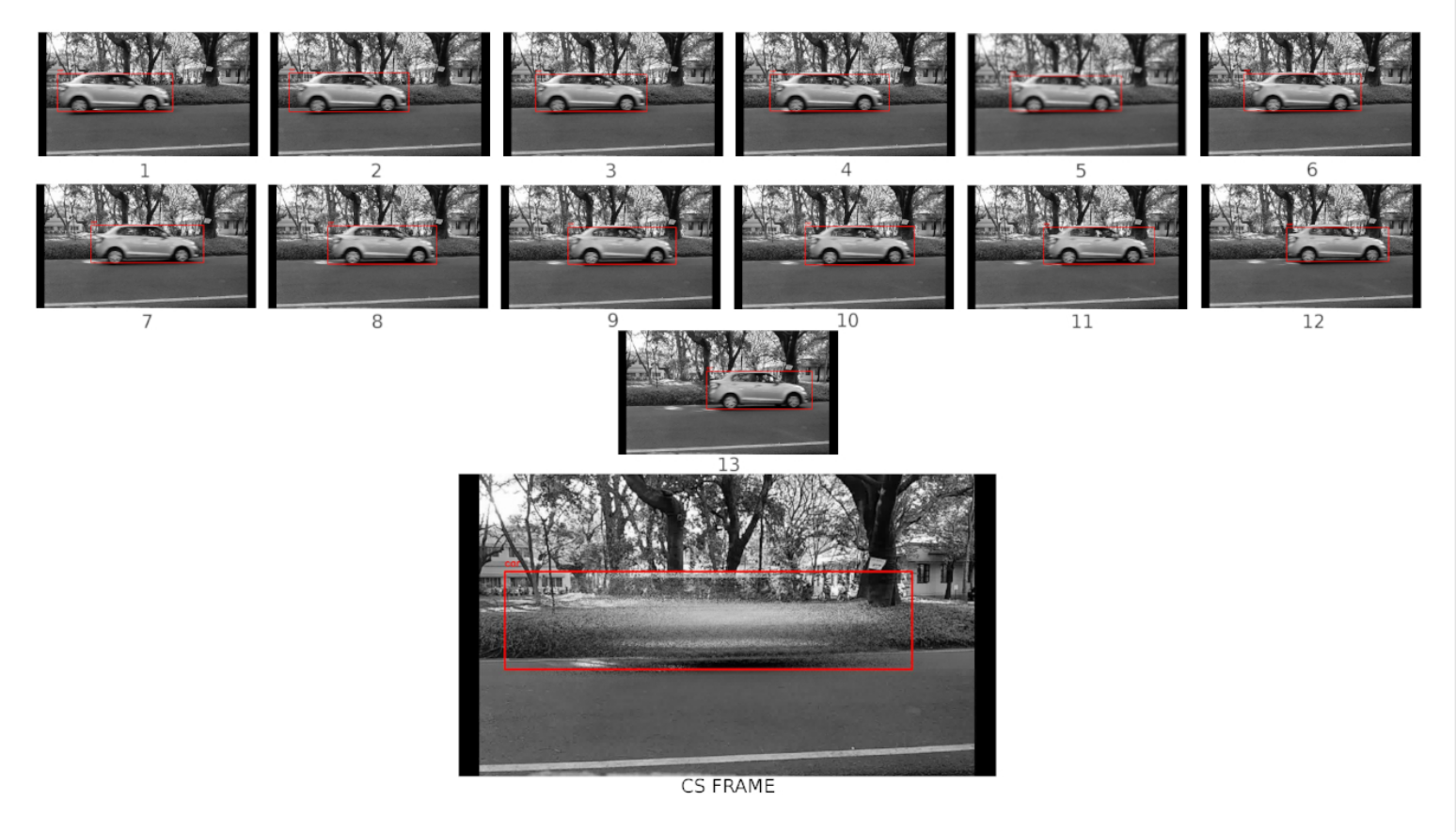}
\caption{Merging boundary box}
\label{fig:fig7}
\end{figure}

\begin{figure}[!h]
\centering
\includegraphics[width= 8cm ]{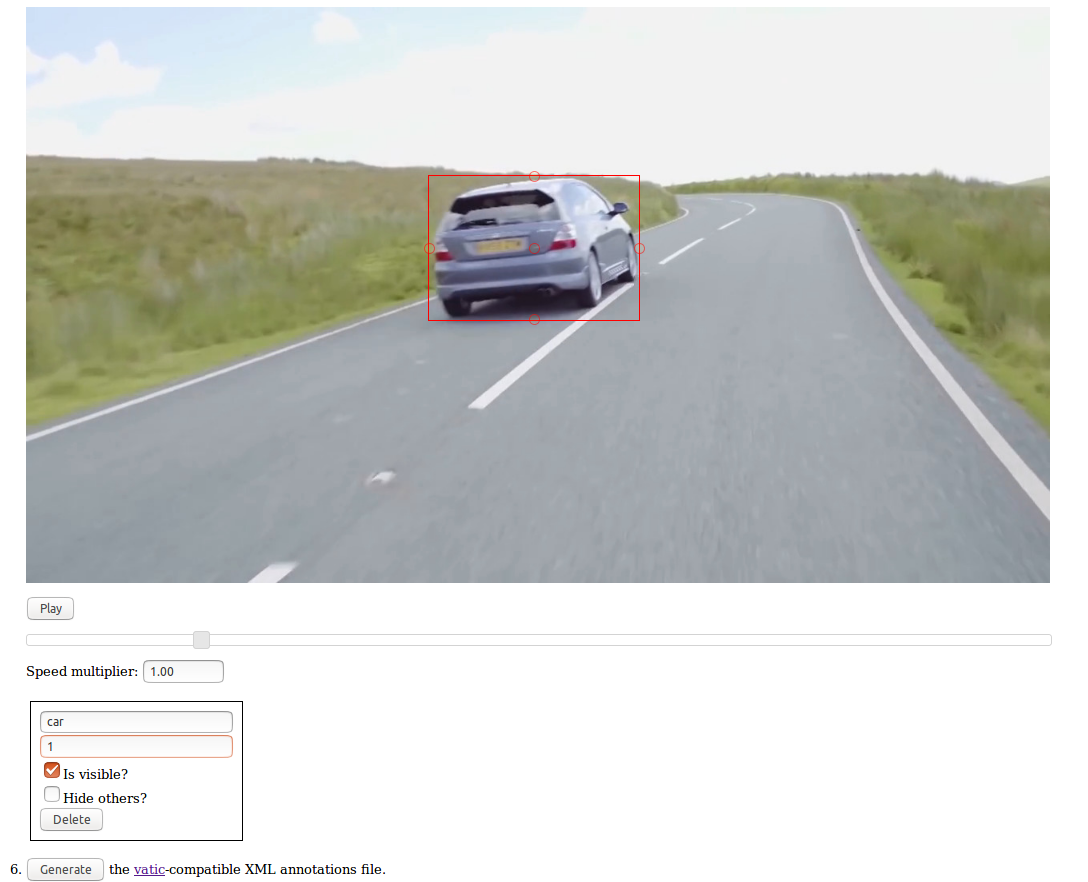}
\caption{Vatic Annotation}
\label{fig:fig8}
\end{figure}

As shown in Fig.\ref{fig:fig5} we used a pre-trained YOLO model to generate the bounding boxes of the individual frames, which were then combined into larger bounding boxes that were used as the ground truth values of the bounding boxes in the CS frames.

We fed these pseudo labeled boxes generated by YOLO with the corresponding frames onto VATIC\cite{vondrick2013efficiently} Where a human corrected the errors Fig.\ref{fig:fig8}. possessed by the Yolo Object detection Model, this kind of pseudo labeling has made it easy to create a large dataset of annotated compressive sensed images.

We wanted less human intervention for the labeling, the pseudo labeling provided an extra hand for the human introspection for labeling. By providing the pseudo labeling the human effort here was reduced and more attention was drawn towards reiterating the boundary box more tightly to fit the objects more precisely and to label the non detected objects in the frame by the Object detection Model Fig.\ref{fig:fig8}.
\begin{figure}[!h]
\centering
\includegraphics[scale=0.6]{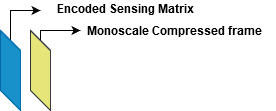} 
\caption{ File Content} 
\label{fig:fig9}
\end{figure}

\item \textbf{File format} \\

Frames: To store compose the dataset in a coherent format, we used NPZ file format by numpy that provides storage of array data using gzip compression. This imageio plugin supports data of any shape and also supports multiple images per file. The line-up composition of these images is as given in the Fig.\ref{fig:fig9}. This kind of setup was realized to elucidate the computation and to automate the process.

Labels : The JSON file holds the $( \text{class ID, frame number and Boundary box information})$ for every frame.

 \end{enumerate}

\end{enumerate}

\section{Discussion and Future work}
To our knowledge, this is the first dataset dedicated to compressive sensing. We aspire a new division of research in these lines which would solve the pressing problems present in computation and space constraints. This data contains a collection of video that is captured by us and a significant portion of video from Youtube8-M dataset. The youtube videos were manually inspected and clipped, so as to avoid any artificial transitions and angle cuts to adhere to the compression theory.

Being the first of its kind, we choose videos with very less density so as to set a baseline for research in the field. The dataset has 284 clips of cars and 91 clips of person of approximately total of 90 minutes of footages put together, with differnt aspect ratios are available in the dataset. In the near future, we tend to release the dataset with different compression rates and bump times in terms of compression with more density videos as well. We also tend to release the segmentation information, pose estimation and other kinds of annotation for this dataset. o as to complete the dataset with all variations and fill the spectrum required for the research and development in this domain. In addition, we set the baseline for a new research topic that would enhance and reduce the computation and hardware expenses to a great extent.

\bibliographystyle{unsrt}  


\end{document}